\documentclass[11pt,a4paper]{article}
\usepackage[hyperref]{acl2017}
\usepackage{times}
\usepackage{latexsym}
\usepackage{amsfonts}
\usepackage{graphicx}
\usepackage{epstopdf}
\usepackage[boxed,linesnumbered,lined]{algorithm2e}
\usepackage{algpseudocode}
\usepackage{multirow}

\usepackage{url}
\usepackage{bm}
\def\argmax{\mathop{\rm argmax}}%
\def\argmin{\mathop{\rm argmin}}%
\def\min{\mathop{\rm min}}%
\newcommand\redbf[1]{\textcolor{red}{\textbf{#1}}}
\newcommand\bluebf[1]{\textcolor{blue}{\textbf{#1}}}
\newcommand\greenbf[1]{\textcolor{green}{\textbf{#1}}}

\aclfinalcopy 
\def\aclpaperid{742} 

\title{Prior Knowledge Integration for Neural Machine Translation \\ using Posterior Regularization}

\author{Jiacheng Zhang$^\dagger$, Yang Liu$^{\dagger\ddagger}$\thanks{\ \ Corresponding author: Yang Liu.} , Huanbo Luan$^\dagger$, Jingfang Xu$^\#$ and Maosong Sun$^{\dagger\ddagger}$ \\
    $^\dagger$State Key Laboratory of Intelligent Technology and Systems  \\
    Tsinghua National Laboratory for Information Science and Technology \\
    Department of Computer Science and Technology, Tsinghua University, Beijing, China \\
    $^\ddagger$Jiangsu Collaborative Innovation Center for Language Competence, Jiangsu, China\\
    $^\#$Sogou Inc., Beijing, China\\
    }

\date{}

\begin{document}
\maketitle
\begin{abstract}
Although neural machine translation has made significant progress recently, how to integrate multiple overlapping, arbitrary prior knowledge sources remains a challenge. In this work, we propose to use posterior regularization to provide a general framework for integrating prior knowledge into neural machine translation. We represent prior knowledge sources as features in a log-linear model, which guides the learning process of the neural translation model. Experiments on Chinese-English translation show that our approach leads to significant improvements. \footnote{The source code is available at \url{https://github.com/Glaceon31/PR4NMT.git}}
\end{abstract}

\section{Introduction}

The past several years have witnessed the rapid development of neural machine translation (NMT) \cite{Sutskever:14,Bahdanau:15}, which aims to model the translation process using neural networks in an end-to-end manner. With the capability of capturing long-distance dependencies due to the gating \cite{Hochreiter:97,Cho:14} and attention \cite{Bahdanau:15} mechanisms, NMT has shown remarkable superiority over conventional statistical machine translation (SMT) across a variety of natural languages \cite{Junczys-Dowmunt:16}.

Despite the apparent success, NMT still suffers from one significant drawback: it is difficult to integrate prior knowledge into neural networks. On one hand, neural networks use {\em continuous} real-valued vectors to represent all language structures involved in the translation process. While these vector representations prove to be capable of capturing translation regularities implicitly \cite{Sutskever:14}, it is hard to interpret each hidden state in neural networks from a linguistic perspective. On the other hand, prior knowledge in machine translation is usually represented in {\em discrete} symbolic forms such as dictionaries and rules \cite{Nirenburg:89} that explicitly encode translation regularities. It is difficult to transform prior knowledge represented in discrete forms to continuous representations required by neural networks.

Therefore, a number of authors have endeavored to integrate prior knowledge into NMT in recent years, either by modifying model architectures \cite{Tu:16,Cohn:16,Tang:16,Feng:16} or by modifying training objectives \cite{Cohn:16,Feng:16,Cheng:16}. For example, to address the over-translation and under-translation problems widely observed in NMT, Tu et al. \shortcite{Tu:16} directly extend standard NMT to model the coverage constraint that each source phrase should be translated into exactly one target phrase \cite{Koehn:03}. Alternatively, Cohn et al. \shortcite{Cohn:16} and Feng et al. \shortcite{Feng:16} propose to control the fertilities of source words by appending additional additive terms to training objectives.

Although these approaches have demonstrated clear benefits of incorporating prior knowledge into NMT, how to combine multiple overlapping, arbitrary prior knowledge sources still remains a major challenge. It is difficult to achieve this end by directly modifying model architectures because neural networks usually impose strong independence assumptions between hidden states. As a result, extending a neural model requires that the interdependence of information sources be modeled explicitly \cite{Tu:16,Tang:16}, making it hard to extend. While this drawback can be partly alleviated by appending additional additive terms to training objectives \cite{Cohn:16,Feng:16}, these terms are restricted to a limited number of simple constraints.

In this work, we propose a general framework for integrating multiple overlapping, arbitrary prior knowledge sources into NMT using {\em posterior regularization} \cite{Ganchev:10}. Our framework is capable of incorporating indirect supervision via posterior distributions of neural translation models. To represent prior knowledge sources as arbitrary real-valued features, we define the posterior distribution as a log-linear model instead of a constrained posterior set \cite{Ganchev:10}. This treatment not only leads to a simpler and more efficient training algorithm but also achieves better translation performance. Experiments show that our approach is able to incorporate a variety of features and achieves significant improvements over posterior regularization using constrained posterior sets on NIST Chinese-English datasets.

\section{Background}

\subsection{Neural Machine Translation}

Given a source sentence $\mathbf{x}=x_1,\dots,x_i, \dots,x_I$ and a target sentence $\mathbf{y}=y_1,\dots,y_j,\dots,y_J$, a neural translation model \cite{Sutskever:14,Bahdanau:15} is usually factorized as a product of word-level translation probabilities:
\begin{eqnarray}
P(\mathbf{y}|\mathbf{x}; \bm{\theta}) = \prod_{j=1}^{J} P(y_j | \mathbf{x}, \mathbf{y}_{<j}; \bm{\theta}),
\end{eqnarray}
where $\bm{\theta}$ is a set of model parameters and $\mathbf{y}_{<j}=y_1,\dots,y_{j-1}$ denotes a partial translation.

The word-level translation probability is defined using a softmax function:
\begin{eqnarray}
P(y_j|\mathbf{x}, \mathbf{y}_{<j}; \bm{\theta}) \propto \exp \Big(f(\mathbf{v}_{y_j}, \mathbf{v}_{\mathbf{x}}, \mathbf{v}_{\mathbf{y}_{<j}}, \bm{\theta}) \Big),
\end{eqnarray}
where $f(\cdot)$ is a non-linear function, $\mathbf{v}_{y_j}$ is a vector representation of the $j$-th target word $y_j$, $\mathbf{v}_{\mathbf{x}}$ is a vector representation of the source sentence $\mathbf{x}$ that encodes the context on the source side, and $\mathbf{v}_{\mathbf{y}_{<j}}$ is a vector representation of the partial translation $\mathbf{y}_{<j}$ that encodes the context on the target side.


Given a training set $\big \{ \langle \mathbf{x}^{(n)}, \mathbf{y}^{(n)} \rangle \big \}_{n=1}^{N}$, the standard training objective is to maximize the log-likelihood of the training set:
\begin{eqnarray}
\hat{\bm{\theta}}_{\mathrm{MLE}} = \argmax_{\bm{\theta}}\Big\{ \mathcal{L}(\theta)  \Big\},
\end{eqnarray}
where
\begin{eqnarray}
\mathcal{L}(\bm{\theta}) = \sum_{n=1}^{N} \log P(\mathbf{y}^{(n)}|\mathbf{x}^{(n)}; \bm{\theta}).
\end{eqnarray}

Although the introduction of vector representations into machine translation has resulted in substantial improvements in terms of translation quality \cite{Junczys-Dowmunt:16}, it is difficult to incorporate prior knowledge represented in discrete symbolic forms into NMT. For example, given a Chinese-English dictionary containing ground-truth translational equivalents such as $\langle${\em baigong}, {\em the White House}$\rangle$, it is non-trivial to leverage the dictionary to guide the learning process of NMT. To address this problem, Tang et al. \shortcite{Tang:16} propose a new architecture called phraseNet on top of RNNsearch \cite{Bahdanau:15} that equips standard NMT with an external memory storing phrase tables.

Another important prior knowledge source is the coverage constraint \cite{Koehn:03}: {\em each source phrase should be translated into exactly one target phrase}. To encode this linguistic intuition into NMT, Tu et al. \shortcite{Tu:16} extend standard NMT with a coverage vector to keep track of the attention history.

While these approaches are capable of incorporating individual prior knowledge sources separately, how to combine multiple overlapping, arbitrary knowledge sources still remains a major challenge. This can be hardly addressed by modifying model architectures because of the lack of interpretability in NMT and the incapability of neural networks in modeling arbitrary knowledge sources. Although modifying training objectives to include additional knowledge sources as additive terms can partially alleviate this problem, these terms have been restricted to a limited number of simple constraints \cite{Cheng:16,Cohn:16,Feng:16} and incapable of combining arbitrary knowledge sources.

Therefore, it is important to develop a new framework for integrating arbitrary prior knowledge sources into NMT.

\subsection{Posterior Regularization}
Ganchev et al. \shortcite{Ganchev:10} propose {\em posterior regularization} for incorporating indirect supervision via constraints on posterior distributions of structured latent-variable models. The basic idea is to penalize the log-likelihood of a neural translation model with the KL divergence between a desired distribution that incorporates prior knowledge and the model posteriors. The posterior regularized likelihood is defined as
\begin{eqnarray}
&& F(\bm{\theta}, \bm{q}) \nonumber \\
&=& \lambda_1 \mathcal{L}(\bm{\theta}) - \nonumber \\
&& \lambda_2 \sum_{n=1}^{N} \min_{q \in \mathcal{Q}} \mathrm{KL}\Big (q(\mathbf{y}) \Big| \Big| P(\mathbf{y}|\mathbf{x}^{(n)}; \bm{\theta}) \Big) , 
\end{eqnarray}
where $\lambda_1$ and $\lambda_2$ are hyper-parameters to balance the preference between likelihood and posterior regularization, $\mathcal{Q}$ is a set of constrained posteriors:
\begin{eqnarray}
\mathcal{Q} = \{ q(\mathbf{y}): \mathbb{E}_{q}[\bm{\phi}(\mathbf{x}, \mathbf{y})] \le \mathbf{b} \},
\end{eqnarray}

where $\bm{\phi}(\mathbf{x}, \mathbf{y})$ is constraint feature and $b$ is the bound of constraint feature expectations. Ganchev et al. \shortcite{Ganchev:10} use constraint features to encode structural bias and define the set of valid distributions with respect to the expectations of constraint features to facilitate inference.

As maximizing $F(\bm{\theta}, \bm{q})$ involves minimizing the KL divergence, Ganchev et al. \shortcite{Ganchev:10} present a minorization-maximization algorithm akin to EM at sentence level:
\begin{eqnarray}
\mathrm{E}: q^{(t+1)} = \argmin_{q}  \mathrm{KL} \Big(q(\mathbf{y})\Big|\Big| P(\mathbf{y}|\mathbf{x}^{(n)}; \bm{\theta}^{(t)})\Big) \nonumber \\
\mathrm{M}: \bm{\theta}^{(t+1)} = \argmax_{\bm{\theta}} \mathbb{E}_{q^{(t+1)}}\Big[ \log P(\mathbf{y}|\mathbf{x}^{(n)}; \bm{\theta}) \Big] \ \nonumber
\end{eqnarray}

However, directly applying posterior regularization to neural machine translation faces a major difficulty: it is hard to specify the hyper-parameter $\mathbf{b}$ to effectively bound the expectation of features, which are usually real-valued in translation \cite{Och:02,Koehn:03,Chiang:05}. For example, the coverage penalty constraint \cite{Wu:16} proves to be an essential feature for controlling the length of a translation in NMT. As the value of coverage penalty varies significantly over different sentences, it is difficult to set an appropriate bound for all sentences on the training data. In addition, the minorization-maximization algorithm involves an additional step to find $q^{(t+1)}$ as compared with standard NMT, which increases training time significantly.

\section{Posterior Regularization for Neural Machine Translation}

\subsection{Modeling}

In this work, we propose to adapt posterior regularization \cite{Ganchev:10} to neural machine translation. The major difference is that we represent the desired distribution as a log-linear model \cite{Och:02} rather than a constrained posterior set as described in \cite{Ganchev:10}:
\begin{eqnarray}
\mathcal{J}(\bm{\theta}, \bm{\gamma}) \quad \quad \quad \quad \quad \quad \quad \quad \quad \quad \quad \quad \quad \ \nonumber \\
= \lambda_1 \mathcal{L}(\bm{\theta}) - \quad \quad \quad \quad \quad \quad \quad \quad \quad \quad \quad \quad \ \nonumber  \\
 \lambda_2 \sum_{n=1}^{N} \mathrm{KL}\Big(Q(\mathbf{y}|\mathbf{x}^{(n)}; \bm{\gamma}) \Big|\Big| P(\mathbf{y}|\mathbf{x}^{(n)}; \bm{\theta})\Big), \label{eq:objective}
\end{eqnarray}
where the desired distribution that encodes prior knowledge is defined as: \footnote{Ideally, the desired distribution $Q$ should be fixed to guide the learning process of $P$. However, it is hard to manually specify the feature weights $\bm{\gamma}$. Therefore, we propose to train both $\bm{\theta}$ and $\bm{\lambda}$ jointly (see Section \ref{sec:training}). We find that joint training results in significant improvements in practice (see Table \ref{table:main}).}
\begin{eqnarray}
Q(\mathbf{y}|\mathbf{x}; \bm{\gamma}) = \frac{\exp \Big( \bm{\gamma} \cdot \bm{\phi}(\mathbf{x}, \mathbf{y}) \Big)}{\sum_{\mathbf{y}'} \exp \Big(\bm{\gamma} \cdot \bm{\phi}(\mathbf{x}, \mathbf{y}') \Big)}.
\end{eqnarray}

As compared to previous work on integrating prior knowledge into NMT \cite{Tu:16,Cohn:16,Tang:16}, our approach provides a general framework for combining arbitrary knowledge sources. This is due to log-linear models that offer sufficient flexibility to represent arbitrary prior knowledge sources as features. We tackle the representation discrepancy problem by associating the $Q$ distribution that encodes {\em discrete} representations of prior knowledge with neural models using {\em continuous} representations learned from data in the KL divergence. Another advantage of our approach is the transparency to model architectures. In principle, our approach can be applied to any neural models for natural language processing.

Our approach also differs from the original version of posterior regularization \cite{Ganchev:10} in the definition of desired distribution. We resort to log-linear models \cite{Och:02} to incorporate features that have proven effective in SMT. Another benefit of using log-linear models is the differentiability of our training objective (see Eq. (\ref{eq:objective})). It is easy to leverage standard stochastic gradient descent algorithms to optimize model parameters (Section \ref{sec:training}).

\subsection{Feature Design}
In this section, we introduce how to design features to encode prior knowledge in the desired distribution.

Note that not all features in SMT can be adopted to our framework. This is because features in SMT are defined on latent structures such as phrase pairs and synchronous CFG rules, which are not accessible to the decoding process of NMT. Fortunately, we can still leverage internal information in neural models that is linguistically meaningful such as the attention matrix $\mathbf{a}$ \cite{Bahdanau:15}.

We will introduce a number of features used in our experiments as follows.

\subsubsection{Bilingual Dictionary}
It is natural to leverage a bilingual dictionary $\mathcal{D}$ to improve neural machine translation. Arthur et al. \shortcite{Arthur:16} propose to incorporate discrete translation lexicons into NMT by using the attention vector to select lexical probabilities on which to be focused.

In our work, for each entry $\langle x, y \rangle \in \mathcal{D}$ in the dictionary, a {\em bilingual dictionary} (BD) feature is defined at the sentence level:
\begin{eqnarray}
\phi_{\mathrm{BD}_{\langle x,y \rangle}}(\mathbf{x}, \mathbf{y}) = \left\{ \begin{array}{ll}
1 & \textrm{if } x \in \mathbf{x} \land y \in \mathbf{y} \\
0 & \textrm{otherwise}
\end{array}
\right. .
\end{eqnarray}
Note that number of bilingual dictionary features depends on the vocabulary of the neural translation model. Entries containing out-of-vocabulary words has to be discarded.

\subsubsection{Phrase Table}
Phrases, which are sequences of consecutive words, are capable of memorizing local context to deal with word ordering within phrases and translation of short idioms, word insertions or deletions \cite{Koehn:03,Chiang:05}. As a result, phrase tables that specify phrase-level correspondences between the source and target languages also prove to be an effective knowledge source in NMT \cite{Tang:16}.

Similar to the bilingual dictionary features, we define a {\em phrase table} (PT) feature for each entry $\langle \tilde{x}, \tilde{y} \rangle$ in a phrase table $\mathcal{P}$:
\begin{eqnarray}
\phi_{\mathrm{PT}_{\langle \tilde{x}, \tilde{y} \rangle}}(\mathbf{x}, \mathbf{y}) = \left\{ \begin{array}{ll}
1 & \textrm{if } \tilde{x} \in \mathbf{x} \land \tilde{y} \in \mathbf{y} \\
0 & \textrm{otherwise}
\end{array}
\right. .
\end{eqnarray}
The number of phrase table features also depends on the vocabulary of the neural translation model.

\subsubsection{Coverage Penalty} \label{sec:cp}
To overcome the over-translation and under-translation problems widely observed in NMT, a number of authors have proposed to model the fertility \cite{Brown:93} and converge constraint \cite{Koehn:03} to improve the adequacy of translation \cite{Tu:16,Cohn:16,Feng:16,Wu:16,Mi:16}.

We follow Wu et al. \shortcite{Wu:16} to define a {\em coverage penalty} (CP) feature to penalize source words with lower sum of attention weights: \footnote{For simplicity, we omit the attention matrix $\mathbf{a}$ in the input of the coverage feature function.}
\begin{eqnarray}
\phi_{\mathrm{CP}}(\mathbf{x}, \mathbf{y}) = \sum_{i=1}^{|\mathbf{x}|} \log \bigg( \min \Big (\sum_{j=1}^{|\mathbf{y}|} \mathbf{a}_{i, j}, 1.0 \Big) \bigg) ,
\label{eq:coverage}
\end{eqnarray}
where $\mathbf{a}_{i, j}$ is the attention probability of the $j$-th target word on the $i$-th source word. Note that the value of coverage penalty feature varies significantly over sentences of different lengths.

\subsubsection{Length Ratio}
Controlling the length of translations is very important in NMT as neural models tend to generate short translations for long sentences, which deteriorates the translation performance of NMT for long sentences as compared with SMT \cite{Shen:16}.

Therefore, we define the {\em length ratio} (LR) feature to encourage the length of a translation to fall in a reasonable range:
\begin{eqnarray}
\phi_{\mathrm{LR}}(\mathbf{x}, \mathbf{y}) = \left\{ \begin{array}{ll}
(\beta |\mathbf{x}|) / |\mathbf{y}| & \textrm{if } \beta|\mathbf{x}| < |\mathbf{y}|\\
|\mathbf{y}| / (\beta |\mathbf{x}|) & \textrm{otherwise}
\end{array}
\right . ,
\end{eqnarray}
where $\beta$ is a hyper-parameter for penalizing too long or too short translations.

For example, to convey the same meaning, an English sentence is usually about 1.2 times longer than a Chinese sentence. As a result, we can set $\beta=1.2$. If the length of a Chinese sentence $|\mathbf{x}|$ is 10 and the length of an English sentence $|\mathbf{y}|$ is 12, then, $\phi_{\mathrm{LR}}(\mathbf{x}, \mathbf{y}) = 1$. If the translation is too long (e.g., $|\mathbf{y}|=100$), then the feature value is 0.12. If the translation is too short (e.g., $|\mathbf{y}|=6$), the feature value is 0.5.

\subsection{Training} \label{sec:training}
In training, our goal is to find a set of model parameters that maximizes the posterior regularized likelihood:
\begin{eqnarray}
\hat{\bm{\theta}}, \hat{\bm{\gamma}} = \argmax_{\bm{\theta}, \bm{\gamma}} \Big\{ \mathcal{J}(\bm{\theta}, \bm{\gamma}) \Big\}.
\end{eqnarray}

Note that unlike the original version of posterior regularization \cite{Ganchev:10} that relies on a minorization-maximization algorithm to optimize model parameters, our training objective is differentiable with respect to model parameters. Therefore, it is easy to use standard stochastic gradient descent algorithms to train our model.

However, a major difficulty in calculating gradients is that the algorithm needs to sum over all candidate translations in an exponential search space for KL divergence. For example, the partial derivative of $\mathcal{J}(\bm{\theta}, \bm{\gamma})$ with respect to $\gamma$ is given by
\begin{eqnarray}
\frac{\partial \mathcal{J}(\bm{\theta}, \bm{\gamma})}{\partial \bm{\gamma}} \quad \quad \quad \quad \quad \quad \quad \quad \quad \quad \quad \quad \ \ \ \nonumber \\
= -\lambda_2 \times \quad \quad \quad \quad \quad \quad \quad \quad \quad \quad \quad \quad \quad \ \ \ \nonumber \\
 \sum_{n=1}^{N} \frac{\partial}{\partial \bm{\gamma}} \mathrm{KL}\Big(Q(\mathbf{y}|\mathbf{x}^{(n)}; \bm{\gamma})\Big|\Big|P(\mathbf{y}|\mathbf{x}^{(n)}; \bm{\theta})\Big).
\end{eqnarray}

The KL divergence is defined as
\begin{eqnarray}
\mathrm{KL}\Big(Q(\mathbf{y}|\mathbf{x}^{(n)}; \bm{\gamma})\Big|\Big|P(\mathbf{y}|\mathbf{x}^{(n)}; \bm{\theta})\Big) \quad \quad \ \ \nonumber \\
= \sum_{\mathbf{y} \in \mathcal{Y}(\mathbf{x}^{(n)})} Q(\mathbf{y}|\mathbf{x}^{(n)}; \bm{\gamma}) \log \frac{Q(\mathbf{y}|\mathbf{x}^{(n)}; \bm{\gamma})}{P(\mathbf{y}|\mathbf{x}^{(n)}; \bm{\theta})},
\end{eqnarray}
where $\mathcal{Y}(\mathbf{x}^{(n)})$ is a set of all possible candidate translations for the source sentence $\mathbf{x}^{(n)}$.

To alleviate this problem, we follow Shen et al. \shortcite{Shen:16} to approximate the full search space $\mathcal{Y}(\mathbf{x}^{(n)})$ with a sampled sub-space $\mathcal{S}(\mathbf{x}^{(n)})$. Therefore, the KL divergence can be approximated as
\begin{eqnarray}
\mathrm{KL}\Big(Q(\mathbf{y}|\mathbf{x}^{(n)}; \bm{\gamma})\Big|\Big|P(\mathbf{y}|\mathbf{x}^{(n)}; \bm{\theta})\Big) \quad \quad \ \ \nonumber \\
\approx \sum_{\mathbf{y} \in \mathcal{S}(\mathbf{x}^{(n)})} \tilde{Q}(\mathbf{y}|\mathbf{x}^{(n)}; \bm{\gamma}) \log \frac{\tilde{Q}(\mathbf{y}|\mathbf{x}^{(n)}; \bm{\gamma})}{\tilde{P}(\mathbf{y}|\mathbf{x}^{(n)}; \bm{\theta})}.
\end{eqnarray}

Note that the $Q$ distribution is also approximated on the sub-space:
\begin{eqnarray}
\tilde{Q}(\mathbf{y}|\mathbf{x}^{(n)}; \bm{\gamma})  \quad \quad \quad \quad \quad \quad \quad \quad \nonumber \\
= \frac{\exp(\bm{\gamma} \cdot \bm{\phi}(\mathbf{x}^{(n)}, \mathbf{y}))}{\sum_{\mathbf{y}' \in \mathcal{S}(\mathbf{x}^{(n)})} \exp(\bm{\gamma} \cdot \bm{\phi}(\mathbf{x}^{(n)}, \mathbf{y}'))}.
\end{eqnarray}

We follow Shen et al. \shortcite{Shen:16} to control the sharpness of approximated neural translation distribution normalized on the sampled sub-space:
\begin{eqnarray}
\tilde{P}(\mathbf{y}|\mathbf{x}^{(n)}; \bm{\theta}) = \frac{P(\mathbf{y}|\mathbf{x}^{(n)}; \bm{\theta})^{\alpha}}{\sum_{\mathbf{y}' \in \mathcal{S}(\mathbf{x}^{(n)})}P(\mathbf{y}'|\mathbf{x}^{(n)}; \bm{\theta})^{\alpha}}.
\end{eqnarray}

\subsection{Search} \label{sec:search}
Given learned model parameters $\hat{\bm{\theta}}$ and $\hat{\bm{\gamma}}$, the decision rule for translating an unseen source sentence $\mathbf{x}$ is given by
\begin{eqnarray}
\hat{\mathbf{y}} = \argmax_{\mathcal{Y}(\mathbf{x})}\Big\{ P(\mathbf{y}|\mathbf{x}; \hat{\bm{\theta}}) \Big\}.
\end{eqnarray}

The search process can be factorized at the word level:
\begin{eqnarray}
\hat{y}_j = \argmax_{y \in \mathcal{V}_y}\Big\{ P(y|\mathbf{x}, \hat{\mathbf{y}}_{<j}; \hat{\bm{\theta}}) \Big\},
\end{eqnarray}
where $\mathcal{V}_y$ is the target language vocabulary.

Although this decision rule shares the same efficiency and simplicity with standard NMT \cite{Bahdanau:15}, it does not involve prior knowledge in decoding. Previous studies reveal that incorporating prior knowledge in decoding also significantly boosts translation performance \cite{Arthur:16,He:16,Wang:16}.

As directly incorporating prior knowledge into the decoding process of NMT depends on both model structure and the locality of features, we resort to a coarse-to-fine approach instead to keep the architecture transparency of our approach. Given a source sentence $\mathbf{x}$ in the test set, we first use the neural translation model $P(\mathbf{y}|\mathbf{x}; \hat{\bm{\theta}})$ to generate a $k$-best list of candidate translation $\mathcal{C}(\mathbf{x})$. Then, the algorithm decides on the most probable candidate translation using the following decision rule:
\begin{eqnarray}
\hat{\mathbf{y}} = \argmax_{\mathbf{y} \in \mathcal{C}(\mathbf{x})}\Big\{ \log P(\mathbf{y}|\mathbf{x}; \hat{\bm{\theta}}) + \hat{\bm{\gamma}} \cdot \bm{\phi}(\mathbf{x}, \mathbf{y}) \Big\}.
\end{eqnarray}

\begin{table*}[!t]
\centering

\begin{tabular}{c|l||c|ccccc|c}
Method &  Feature & MT02 & MT03 & MT04 & MT05 & MT06 & MT08 & All \\
\hline \hline
\textproc{RNNsearch}  & N/A & 33.45 & 30.93 & 32.57 & 29.86 & 29.03 & 21.85 & 29.11 \\
\hline \hline
\textproc{CPR}  & N/A & 33.84 & 31.18 & 33.26 & 30.67 & 29.63 & 22.38 & 29.72 \\
\hline
\multirow{5}{*}{\textproc{PostReg}}  & BD & 34.65 & 31.53 & 33.82 & 30.66 & 29.81 & 22.55 & 29.97 \\
  & PT & 34.56 & 31.32 & 33.89 & 30.70 & 29.84 & 22.62 & 29.99 \\
  & LR & 34.39 & 31.41 & 34.19 & 30.80 & 29.82 & 22.85 & 30.14 \\
  & BD+PT & 34.66 & 32.05 & 34.54 & 31.22 & 30.70 & 22.84 & 30.60 \\
  & BD+PT+LR & 34.37 & 31.42 & 34.18 & 30.99 & 29.90 & 22.87 & 30.20 \\
\hline \hline
\multirow{7}{*}{{\em this work}} & BD & \textbf{36.61} & 33.47 & 36.04 & 32.96 & 32.46 & \textbf{24.78} & 32.27 \\
& PT & 35.07 & 32.11 & 34.73 & 31.84 & 30.82 & 23.23 & 30.86 \\
& CP & 34.68 & 31.99 & 34.67 & 31.37 & 30.80 & 23.34 & 30.76 \\
& LR & 34.57 & 31.89 & 34.95 & 31.80 & 31.43 & 23.75 & 31.12 \\
& BD+PT & 36.30 & \textbf{33.83} & 36.02 & 32.98 & 32.53 & 24.54 & 32.29 \\
& BD+PT+CP & 36.11 & 33.64 & 36.36 & \textbf{33.11} & 32.53 & 24.57 & 32.39 \\
& BD+PT+CP+LR & 36.10 & 33.64 & \textbf{36.48} & 33.08 & \textbf{32.90} & 24.63 & \textbf{32.51}
\end{tabular}
\caption{Comparison of BLEU scores on the Chinese-English datasets. \textproc{RNNSearch} is an attention-based neural machine translation model \cite{Bahdanau:15} that does not incorporate prior knowledge. \textproc{CPR} extends \textproc{RNNsearch} by introducing coverage penalty refinement (Eq. (\ref{eq:coverage})) in decoding. \textproc{PostReg} extends \textproc{RNNsearch} with posterior regularization \cite{Ganchev:10}, which uses constraint features to represent prior knowledge and a constrained posterior set to denote the desired distribution. Note that \textproc{PostReg} cannot use the CP feature (Section \ref{sec:cp}) because it is hard to bound the feature value appropriately. On top of \textproc{RNNsearch}, our approach also exploits posterior regularization to incorporate prior knowledge but uses a log-linear model to denote the desired distribution. All results of {\em this work} are significantly better than \textproc{RNNsearch} ($p < 0.01$).} \label{table:main}
\end{table*}

\section{Experiments}

\subsection{Setup}

We evaluate our approach on Chinese-English translation. The evaluation metric is case-insensitive BLEU calculated by the {\em multi-bleu.perl} script. Our training set\footnote{The training set includes LDC2002E18, LDC2003E07, LDC2003E14, part of LDC2004T07, LDC2004T08 and LDC2005T06.} consists of 1.25M sentence pairs with 27.9M Chinese words and 34.5M English words. We use the NIST 2002 dataset as validation set and the NIST 2003, 2004, 2005, 2006, 2008 datasets as test sets.

In the experiments, we compare our approach with the following two baseline approaches:

\begin{enumerate}
\item \textproc{RNNsearch} \cite{Bahdanau:15}: a standard attention-based neural machine translation model,
\item \textproc{CPR} \cite{Wu:16}: extending \textproc{RNNsearch} by introducing coverage penalty refinement (Eq. (\ref{eq:coverage})) in decoding,
\item \textproc{PostReg} \cite{Ganchev:10}: extending \textproc{RNNsearch} with posterior regularization using constrained posterior set.
\end{enumerate}

For \textproc{RNNsearch}, we use an in-house attention-based NMT system that achieves comparable translation performance with \textproc{GroundHog} \cite{Bahdanau:15}, which serves as a baseline approach in our experiments. We limit vocabulary size to 30K for both languages. The word embedding dimension is set to 620. The dimension of hidden layer is set to 1,000. In training, the batch size is set to 80. We use the AdaDelta algorithm \cite{Zeiler:12} for optimizing model parameters. In decoding, the beam size is set to 10.

For \textproc{CPR}, we simply follow Wu et al. \shortcite{Wu:16} to incorporate the coverage penalty into the beam search algorithm of \textproc{RNNsearch}.

For \textproc{PostReg}, we adapt the original version of posterior regularization \cite{Ganchev:10} to NMT on top of \textproc{RNNsearch}. Following Ganchev et al. \shortcite{Ganchev:10}, we use a ten-step projected gradient descent algorithm to search for an approximate desired distribution in the E step and a one-step gradient descent for the M step.

Our approach extends \textproc{RNNsearch} by incorporating prior knowledge. For each source sentence, we sample 80 candidate translations to approximate the $\tilde{P}$ and $\tilde{Q}$ distributions. The hyper-parameter $\alpha$ is set to 0.2. The batch size is 1. The hyper-parameters $\lambda_1$ and $\lambda_2$ are set to $8\times 10^{-5}$ and $2.5 \times 10^{-4}$. Note that they not only balance the preference between likelihood and posterior regularization, but also control the values of gradients to fall in a reasonable range for optimization.

\begin{table*}[!t]
\centering
\begin{tabular}{l|c||c|ccccc|c}
Feature & Rerank & MT02 & MT03 & MT04 & MT05 & MT06 & MT08 & All \\
\hline \hline
\multirow{2}{*}{BD} & w/o & 36.06 & 32.99 & 35.62 & 32.59 & 32.13 & 24.36 & 31.87 \\
& w/ & \textbf{36.61} & 33.47 & 36.04 & 32.96 & 32.46 & \textbf{24.78} & 32.27 \\
\hline
\multirow{2}{*}{PT} & w/o & 34.98 & 32.01 & 34.71 & 31.77 & 30.77 & 23.20 & 30.81 \\
& w/ & 35.07 & 32.11 & 34.73 & 31.84 & 30.82 & 23.23 & 30.86 \\
\hline
\multirow{2}{*}{CP} & w/o & 34.68 & 31.99 & 34.67 & 31.37 & 30.80 & 23.34 & 30.76 \\
& w/ & 34.68 & 31.99 & 34.67 & 31.37 & 30.80 & 23.34 & 30.76 \\
\hline
\multirow{2}{*}{LR} & w/o & 34.60 & 31.89 & 34.79 & 31.72 & 31.39 & 23.63 & 31.03 \\
& w/ & 34.57 & 31.89 & 34.95 & 31.80 & 31.43 & 23.75 & 31.12 \\
\hline
\multirow{2}{*}{BD+PT} & w/o & 35.76 & 33.27 & 35.64 & 32.47 & 32.03 & 24.17 & 31.83 \\
& w/ & 36.30 & \textbf{33.83} & 36.02 & 32.98 & 32.53 & 24.54 & 32.29 \\
\hline
\multirow{2}{*}{BD+PT+CP} & w/o & 35.71 & 33.15 & 35.81 & 32.52 & 32.16 & 24.11 & 31.89 \\
& w/ & 36.11 & 33.64 & 36.36 & \textbf{33.11} & 32.53 & 24.57 & 32.39 \\
\hline
\multirow{2}{*}{BD+PT+CP+LR} & w/o & 36.06 & 33.01 & 35.86 & 32.70 & 32.24 & 24.27 & 31.96 \\
& w/ & 36.10& 33.64 & \textbf{36.48} & 33.08 & \textbf{32.90} & 24.63 & \textbf{32.51} \\
\end{tabular}
\caption{Effect of reranking on translation quality.} \label{table:rerank}
\end{table*}

\begin{table*}[!t]
\centering
\begin{tabular}{l|p{1.5\columnwidth}}
\hline
Source & {\em lijing liang tian yu bingxue de \bluebf{fenzhan} , 31ri shenye 23 shi 50 fen , shanghai jichang jituan \redbf{yuangong} \greenbf{yinglai} le 2004nian de zuihou yige hangban .} \\
\hline
Reference & after \bluebf{fighting} with ice and snow for two days , \redbf{staff} members of shanghai airport group \greenbf{welcomed} the last flight of 2004 at 23 : 50pm on the 31st .\\
\hline
\textproc{RNNsearch} & after a two - day and two - day journey , the team of shanghai 's airport in shanghai has ushered in the last flight in 2004 . \\
\hline
+ BD & after two days and nights \bluebf{fighting} with ice and snow , the shanghai airport group 's \redbf{staff} \greenbf{welcomed} the last flight in 2004 . \\
\hline
\hline
Source & {\em suiran tonghuopengzhang \bluebf{weilai ji ge yue} reng jiang weizhi zai \redbf{baifenzhier yishang} , buguo niandi zhiqian keneng jiangdi .} \\
\hline
Reference & although inflation will remain \redbf{above 2 \%} \bluebf{for the coming few months} , it may decline by the end of the year .\\
\hline
\textproc{RNNsearch} & although inflation has been maintained for more than two months from the year before the end of the year , it may be lower . \\
\hline
+ PT & although inflation will remain at \redbf{more than 2 percent} \bluebf{in the next few months} , it may be lowered before the end of the year . \\
\hline
\hline
Source & {\em qian ji tian ta ganggang \bluebf{chuyuan} , jintian \redbf{jianchi} lai yu lao pengyou \redbf{daobie} .} \\
\hline
Reference & just \bluebf{discharged from the hospital} a few days ago , he \redbf{insisted on} coming to \redbf{say farewell} to his old friend today .\\
\hline
\textproc{RNNsearch} & during the previous few days , he had just been given treatment to the old friends .\\
\hline
+ CP & during the previous few days , he had just been \bluebf{discharged from the hospital} , and he \redbf{insisted on goodbye} to his old friend today . \\
\hline
\hline
Source & {\em ( guoji ) yiselie fuzongli fouren jihua kuojian \redbf{gelan gaodi} dingjudian} \\
\hline
Reference &  ( international ) israeli deputy prime minister denied plans to expand \redbf{golan heights} settlements\\
\hline
\textproc{RNNsearch} &  ( world ) israeli deputy prime minister denies the plan to expand \redbf{the golan heights} in \redbf{the golan heights}\\
\hline
+ LR & ( international ) israeli deputy prime minister denies planning to expand \redbf{golan heights} \\
\hline
\end{tabular}
\caption{Example translations that demonstrate the effect of adding features.} \label{table:example}
\end{table*}

We construct bilingual dictionary and phrase table in an automatic way. First, we run the statistical machine translation system \textproc{Moses} \cite{Koehn:07} to obtain probabilistic bilingual dictionary and phrase table. For the bilingual dictionary, we retain entries with probabilities higher than 0.1 in both source-to-target and target-to-source directions. For phrase table, we first remove phrase pairs that occur less than 10 times and then retain entries with probabilities higher than 0.5 in both directions. As a result, both bilingual dictionary and phrase table contain high-quality translation correspondences. We estimate the length ratio on Chinese-English data and set the hyper-parameter $\beta$ to 1.236.

By default, both \textproc{PostReg} and our approach use reranking to search for the most probable translations (Section \ref{sec:search}).

\subsection{Main Results}

Table \ref{table:main} shows the BLEU scores obtained by \textproc{RNNsearch}, \textproc{PostReg}, and our approach on the Chinese-English datasets.

We find \textproc{PostReg} achieves significant improvements over \textproc{RNNsearch} by adding features that encode prior knowledge. The most effective single feature for \textproc{PostReg} seems to be the length ratio (LR) feature, suggesting that it is important for NMT to control the length of translation to improve translation quality. Note that \textproc{PostReg} is unable to include the coverage penalty (CP) feature because the feature value varies significantly over different sentences. It is hard to specify an appropriate bound $\mathbf{b}$ for constraining the expected feature value. We observe that a loose bound often makes the training process very unstable and fail to converge. Combining features obtains further modest improvements.

Our approach outperforms both \textproc{RNNSearch} and \textproc{PostReg} significantly. The bilingual dictionary (BD) feature turns out to make the most contribution. Compared with \textproc{CPR} that imposes coverage penalty during decoding, our approach that using a single CP feature obtains a significant improvement (i.e., 30.76 over 29.72), suggesting that incorporating prior knowledge sources in modeling might be more beneficial than in decoding.

We find that combining features only results in modest improvements for our approach. One possible reason is that the bilingual dictionary and phrase table features overlap on single word pairs.

\subsection{Effect of Reranking}

Table \ref{table:rerank} shows the effect of reranking on translation quality. We find that using prior knowledge features to rescore the $k$-best list produced by the neural translation model usually leads to improvements. This finding confirms that adding prior knowledge is beneficial for NMT, either in the training or decoding process.

\subsection{Training Speed}

Initialized with the best \textproc{RNNSearch} model trained for 300K iterations, our model converges after about 100K iterations. For each iteration, our approach is 1.5 times slower than \textproc{RNNSearch}. On a single GPU device Tesla M40, it takes four days to train the \textproc{RNNSearch} model and three extra days to train our model.

\subsection{Example Translations}

Table \ref{table:example} gives four examples to demonstrate the benefits of adding features.

In the first example, source words ``fenzhan'' ({\em fighting}), ``yuangong'' ({\em staff}), and ``yinglai'' ({\em welcomed}) are untranslated in the output of \textproc{RNNsearch}. Adding the bilingual dictionary (BD) feature encourages the model to translate these words if they occur in the dictionary.

In the second example, while \textproc{RNNsearch} fails to capture phrase cohesion, adding the phrase table (PT) feature is beneficial for translating short idioms, word insertions or deletions that are sensitive to local context.

In the third example, \textproc{RNNsearch} tends to omit many source content words such as ``chuyuan'' ({\em discharged from the hospital}), ``jianchi'' ({\em insisted on}), and ``daobie'' ({\em say farewell}). The coverage penalty (CP) feature helps to alleviate the word omission problem.

In the fourth example, the translation produced by \textproc{RNNsearch} is too long and ``the golan heights'' occurs twice. The length ratio (LR) feature is capable of controlling the sentence length in a reasonable range.

\section{Related Work}

Our work is directly inspired by posterior regularization \cite{Ganchev:10}. The major difference is that we use a log-linear model to represent the desired distribution rather than a constrained posterior set. Using log-linear models not only enables our approach to incorporate arbitrary knowledge sources as real-valued features, but also is differentiable to be jointly trained with neural translation models efficiently.

Our work is closely related to recent work on injecting prior knowledge into NMT \cite{Arthur:16,Tu:16,Cohn:16,Tang:16,Feng:16,Wang:16}. The major difference is that our approach aims to provide a general framework for incorporating arbitrary prior knowledge sources while keeping the neural translation model unchanged.

He et al. \shortcite{He:16} also propose to combine the strengths of neural networks on learning representations and log-linear models on encoding prior knowledge. But they treat neural translation models as a feature in the log-linear model. In contrast, we connect the two models via KL divergence to keep the transparency of our approach to model architectures. This enables our approach to be easily applied to other neural models in NLP.

\section{Conclusion}

We have presented a general framework for incorporating prior knowledge into end-to-end neural machine translation based on posterior regularization \cite{Ganchev:10}. The basic idea is to guide NMT models towards desired behavior using a log-linear model that encodes prior knowledge.  Experiments show that incorporating prior knowledge leads to significant improvements over both standard NMT and posterior regularization using constrained posterior sets.


\section*{Acknowledgments}

We thank Shiqi Shen for useful discussions and anonymous reviewers for insightful comments. This work is supported by the National Natural Science Foundation of China (No.61432013), the 973 Program (2014CB340501), and the National Natural Science Foundation of China (No.61522204). This research is also supported by Sogou Inc. and the Singapore National Research Foundation under its International Research Centre@Singapore Funding Initiative and administered by the IDM Programme. 

\bibliography{acl2017_zjc}
\bibliographystyle{acl_natbib}

\end{document}